\documentclass[conference]{IEEEtran}
\IEEEoverridecommandlockouts
\usepackage{cite}
\usepackage{amsmath,amssymb,amsfonts}
\usepackage{algorithmic}
\usepackage{graphicx}
\usepackage{textcomp}
\usepackage{xcolor}
\usepackage{multirow}
\usepackage{array}
\usepackage[numbers,sort&compress]{natbib}
\usepackage{CJKutf8}
\usepackage{lipsum}

\begin{document}

\title{
SLK-NER: Exploiting Second-order Lexicon Knowledge for Chinese NER
}

\author{
\IEEEauthorblockN{Dou Hu\IEEEauthorrefmark{1} and  Lingwei Wei \IEEEauthorrefmark{2} \IEEEauthorrefmark{3}   
} 
  
\IEEEauthorblockA{\IEEEauthorrefmark{1}National Computer System Engineering Research Institute of China, Beijing, China } 
\IEEEauthorblockA{\IEEEauthorrefmark{2}School of Cyber Security, University of Chinese Academy of Sciences, Beijing, China} 
\IEEEauthorblockA{\IEEEauthorrefmark{3}Institute of Information Engineering, Chinese Academy of Sciences, Beijing, China \\
hudou18@mails.ucas.edu.cn, weilingwei@iie.ac.cn} 
}



\maketitle
\newcommand\blfootnote[1]{%
\begingroup 
\renewcommand\thefootnote{}\footnote{#1}%
\addtocounter{footnote}{-1}%
\endgroup 
}

\begin{abstract}
    Although character-based models using lexicon have achieved promising results 
    for Chinese named entity recognition (NER) task,
    some lexical words would introduce
    erroneous information due to wrongly matched words. 
    Existing researches proposed many strategies to 
    integrate lexicon knowledge.
    However, they performed with simple first-order lexicon knowledge,
    which provided insufficient word information and still faced the challenge of matched word boundary conflicts;
    or explored the lexicon knowledge with graph 
     where higher-order information introducing negative words may disturb the identification.
    
    To alleviate the above limitations, 
    we present new insight into second-order lexicon knowledge (SLK) of each character in the sentence
    to provide more lexical word information including semantic and word boundary features.
    Based on these, we propose a SLK-based model 
    with a novel strategy to integrate the above lexicon knowledge.
    The proposed model can exploit more discernible lexical words information
    with the help of global context.
    %
    %
    Experimental results on three public datasets demonstrate the validity of SLK.
    The proposed model achieves more excellent performance than the state-of-the-art comparison methods.
    
\end{abstract}


\blfootnote{DOI reference number: 10.18293/SEKE2020-153.}

\section{Introduction}

Named Entity Recognition (NER) aims to locate and classify named entities into predefined entity categories in the corpus,
which is a fundamental task for various downstream applications 
 such as information retrieval \cite{guo2009named},
question answering \cite{diefenbach2018core}, machine translation \cite{dandapat2016improved}, etc.
Word boundaries in Chinese are ambiguities and word segmentation errors 
have a negative impact on identifying Name Entity (NE) \cite{peng2015named},
which would make Chinese NER more difficult to identify. 
Explicit discussions have approved that character-based taggers can outperform word-based counterparts \cite{zhang2018chinese}.

\begin{figure}
    \centerline{\includegraphics[width=0.45\textwidth]{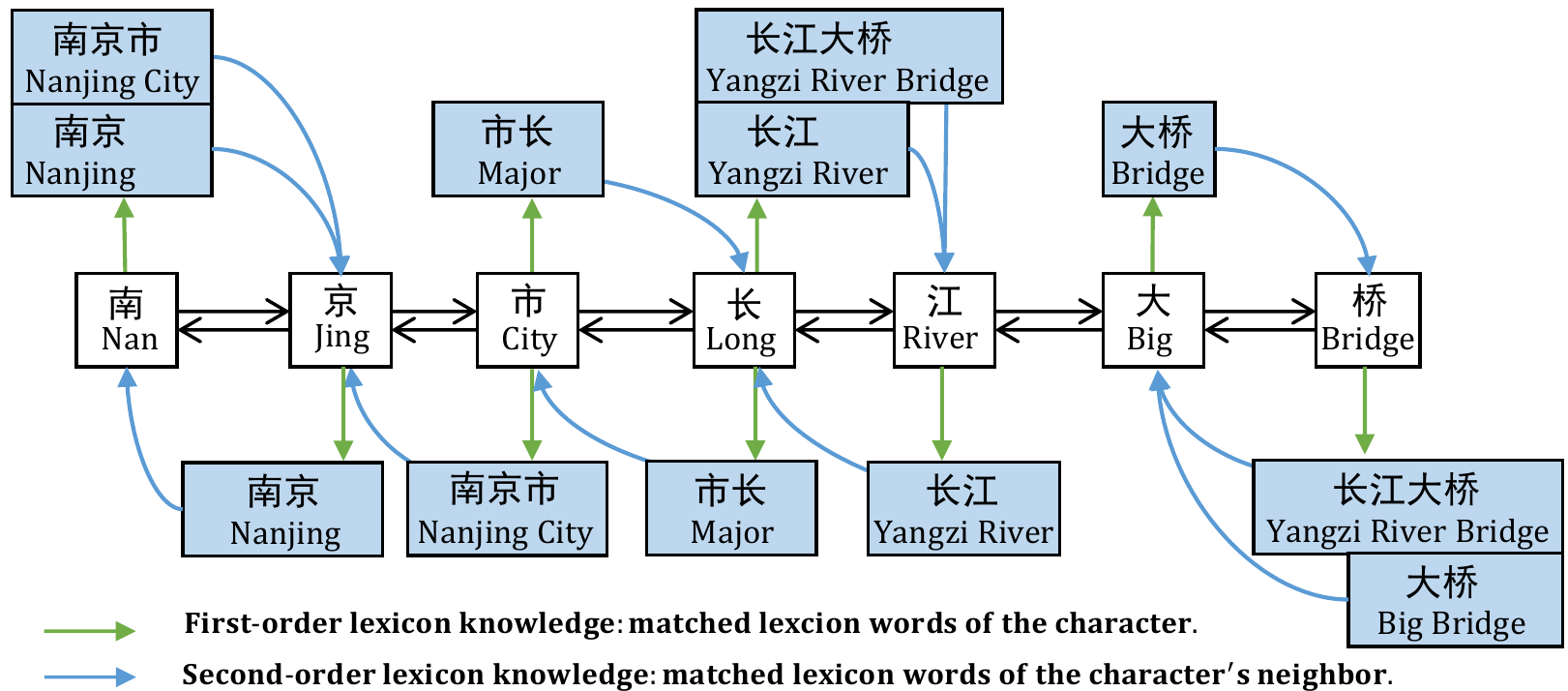}}
    \caption{An example of a word character lattice. 
    The top is that predicting the label uses from left to right sequence and 
    the bottom is using from right to left sequence. 
    The green arrow line represents 
    that the character (start) can match with the lexical word (end).
    The blue arrow line represents the lexical word (start) information would be integrated into the character (end).
    } \label{example}
    \end{figure}

Because entity boundaries usually coincide with some word boundaries,
integrating external lexicon knowledge into character-based models
has attracted research attention \cite{zhang2018chinese}. 
Although lexicon can be useful, in practice the lexical words may introduce
erroneous information and suffer from word boundary conflicts, which easily lead to wrongly matched entities
and limit system the performance \cite{chiu2016named}.
To address the above issues, many sequence-based efforts have been devoted to incorporated lexicon knowledge into sentences
  \cite{gui2019cnn,liu2019encoding}.

\begin{CJK*}{UTF8}{gbsn}
However, these strategies explore simple \textit{first-order lexicon knowledge}(FLK) of each character
as shown in the green arrow line in Fig.\ref{example}. 
FLK only contains the lexical features of the characters itself,
which cannot offer adequate word information.
For example, the character ``京(Jing)" only introduces ``南京(Nanjing)" based on FLK.
The wrongly matched word information would misidentify as ``南京(Nanjing)" instead of ``南京市(Nanjing City)". 
As a result, they continue to suffer from boundary conflicts between 
potential words being incorporating in the lexicon.
The conflict caused by this deficiency mainly comes from the middle of the named entity,
such as ``大(Big)" and ``江(River)" in ``长江大桥(Yangtze River Bridge)". 
\end{CJK*}

Recently, some models attempted to aggregate rich higher-order lexicon knowledge, such as 
graph structure \cite{gui2019lexicon,sui2019leverage,ding2019neural}.
This higher-order information probably introduces irrelevant words with the character, limiting the performance to some extent.
In addition, the existence of shortcut paths may cause the model degeneration into a partially word-based model,
 which would suffer from segmentation errors.  

\begin{CJK*}{UTF8}{gbsn}
To address the above issue, 
we introduce the \textit{second-order lexicon knowledge} (SLK) to each character in the input sentence,
that is the neighbor's lexicon knowledge of the character, 
as elaborated in Fig.\ref{example} with the blue arrow lines.
The SLK of ``京(Jing)" contains both ``南京市(Nanjing City)" and ``南京(Nanjing)" from its left neighbor ``南(Nan)", 
and ``南京市(Nanjing City)" from its right neighbor ``市(City)".
With regard to global semantics of the sentence,
``南京市(Nanjing City)" is more likely to be the named entity than ``南京(Nanjing)" 
due to higher semantic similarity of ``南京市(Nanjing City)". 
Similarly, the SLK of ``江(River)" is the potential words ``长江大桥(Yangtze River Bridge)" and ``长江(Yangtze River)",
and the SLK of ``大(Big)" is ``长江大桥(Yangtze River Bridge)" and ``大桥(Big Bridge)".
By synthesizing global considerations, these lexicon knowledge guides 
the character subsequence ``长江大桥(Yangtze River Bridge)" to be recognized as the named entity.  
\end{CJK*}









To take advantage of this insight, we proposed a SLK-based model with a novel strategy named SLK-NER,
to integrate more informative lexicon words into the character-based model.
Specifically, we assign SLK to each character and ensure no shortcut paths between characters. 
Furthermore, we 
utilize global contextual information to fuse the lexicon knowledge via attention mechanism. 
The model enables capture more useful lexical word features automatically
and relieves the word boundary conflicts problem for better Chinese NER performance.

The main contributions can be summarized as follows:
\begin{itemize}
    \item \textbf{Insight.} 
We present a new insight about second-order lexicon knowledge (SLK) of the character. 
    SLK can provide sufficient lexicon knowledge into characters in sentences
    and is capable of relieving the challenge of word boundary conflicts.
    \item \textbf{Method. }
    To properly leverage SLK, we propose a Chinese NER model named SLK-NER with a novel strategy
    to integrate lexicon knowledge into the character-based model.
    SLK-NER can enable to capture more beneficial word features 
    with the help of global context information via attention mechanism.
    
    \item \textbf{Evaluation.}
    Experimental results demonstrate the efficiency of SLK 
    and our model significantly outperforms pervious methods,
     achieving state-of-the-art over three public Chinese NER datasets. 
    The source code and dataset are available\footnote{https://github.com/zerohd4869/SLK-NER}.
\end{itemize}

\section{Related Works}

Early character-based methods for NER considered few word information in character sequence \cite{peng2015named,devlin-etal-2019-bert,huang2015bidirectional,zhu2019can}. 
To tackle this limitation, many works generally use lexicon as extra word information for Chinese NER.

\subsection{Sequence-based Methods}
Zhang et al. \cite{zhang2018chinese} introduced a lattice LSTM to model all potential words matching 
a sentence to exploit explicit word information and achieved state-of-the-art results.
Lattice LSTM enlightened various approaches for the useage of lexicon knowledge.
Chain-structured LSTM \cite{liu2019encoding}  integrated word boundary features into input character vector via four strategies.
Gui et al. \cite{gui2019cnn} extended rethinking mechanism to relieve word boundary conflicts.


\begin{figure}[t]
    \centerline{\includegraphics[width=0.5\textwidth]{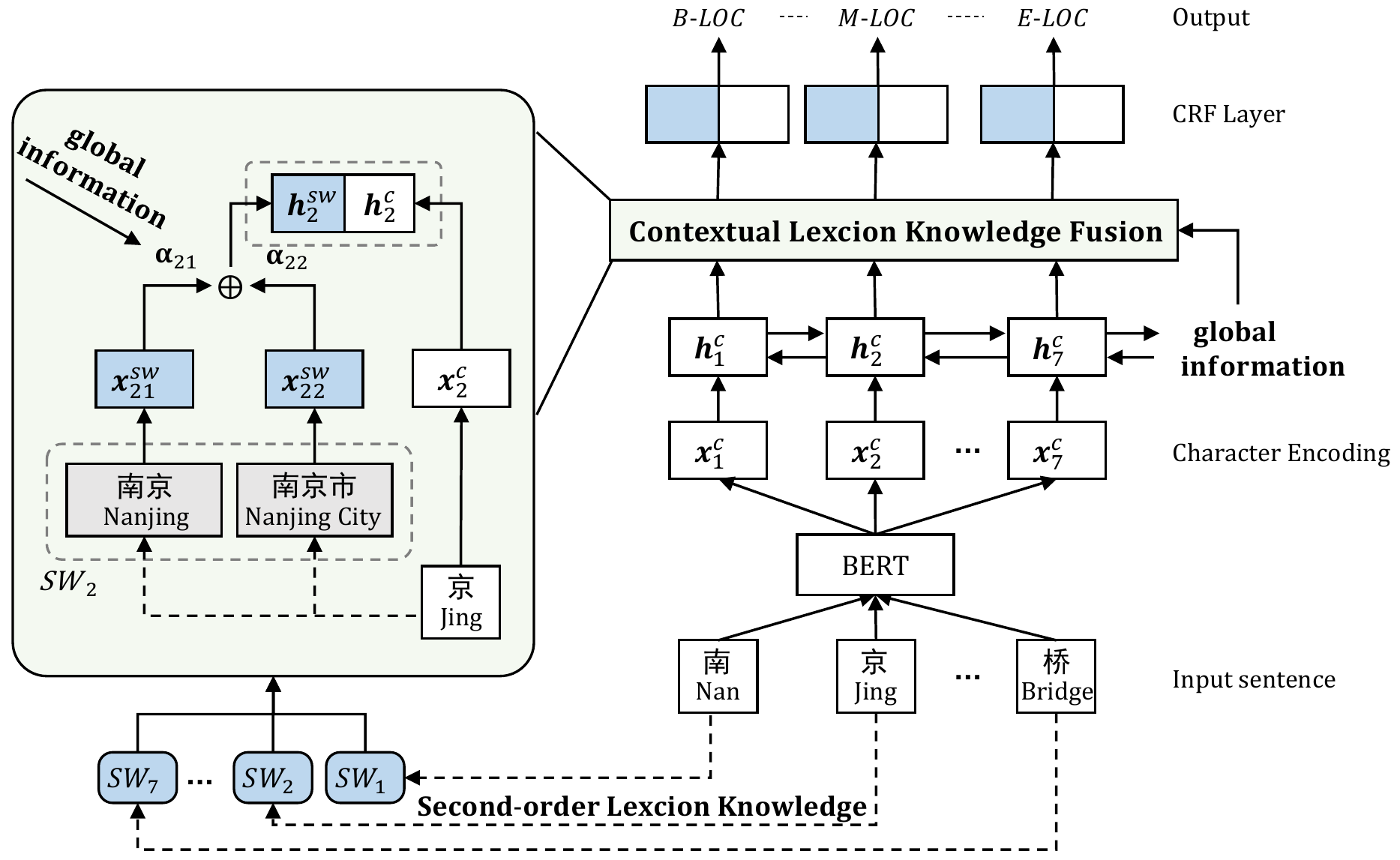}}
    \caption{The whole architecture of SLK-NER. 
   It is comprised of character encoding layer, lexicon knowledge encoding layer, 
contextual lexicon knowledge fusion and a CRF decoding layer.
    } \label{overview}
    \end{figure}

\subsection{Graph-based Methods}
With the development of graph, there are some studies improved by graph neural networks.
For instance, Gui et al. \cite{gui2019lexicon} proposed a GNN-based method to explore multiple graph-based interactions among characters, potential words,
and the whole-sentence semantics and effectively alleviated the word ambiguity.
Sui et al. \cite{sui2019leverage} proposed a collaborative graph network to assign both self-matched and the nearest contextual lexical words. 
Ding et al. \cite{ding2019neural} proposed a multi-digraph structure to learn the contextual information
of the characters and the lexicon.



\section{Method}

\subsection{Overview}

The overall architecture of our proposed model is illustrated in Fig.\ref{overview}.
First, we encode character-based sentences 
to explicitly capture the contextual features of the sentence via character encoding layer.
Second, to integrate more lexicon knowledge, we construct the second-order lexicon knowledge (SLK) for each character.
Third, a fusion layer with the global attention information is used for fusing different SLK 
to alleviate the impact of word boundary conflicts.
Finally, a standard CRF model \cite{lafferty2001conditional} is 
employed for decoding labels. 

Formally, we denote an input sentence as $s = \{ c_{1}, c_{2}..., c_{n} \}$,
where $c_i$ means the $i$th character. 
The lexicon $\mathcal{D}$ is the same as \cite{zhang2018chinese}, which is built by 
using automatically segmented large raw text.
For $i$th character, we use $\overrightarrow{\mathcal{FW}}_i$ to denote a set of words obtained by 
matching all possible forward subsequences in lexicon $\mathcal{D}$ \cite{liu2019encoding}.  
Similarly, we use ${\overleftarrow{\mathcal{FW}}_i}$ to denote the words 
for $i$th character in backward process.
The knowledge involved in these sets represents the FLK corresponding to the $i$th character,
i.e., $\mathcal{FW}_i = {\overrightarrow{\mathcal{FW}}_i} \cup {\overleftarrow{\mathcal{FW}}_i}$.
Based on FLK, 
SLK of $i$th character can be defined as:
\begin{equation}
    \mathcal{SW}_i = \overrightarrow{\mathcal{FW}}_{i-1}  \cup  \overleftarrow{\mathcal{FW}}_{i+1}
    , i \in [1,n].
\end{equation}
\begin{CJK*}{UTF8}{gbsn}
As the example shows in Fig.\ref{overview}, SLK of the character ``京(Jing)" is 
the word set including ``南京(Nanjing)" and ``南京市(Nanjing City)".
\end{CJK*}
SLK can mitigate the negative impact of word boundary conflicts. Therefore, we utilize SLK
in our proposed model.

\subsection{Character Encoding Layer}
Given the sentence $s$, 
a pre-trained model BERT \cite{devlin-etal-2019-bert} 
encodes each character $c_i$ in the sentence to a vector. 
\begin{equation}
    \textbf{x}^c_i=BERT(c_i).
\end{equation}
To capture more contextual information, 
we apply bi-directional Gate Recurrent Unit: 
\begin{equation}
    {\textbf{h}^c_i}={GRU}(\textbf{x}^c_i),i \in [1,n]. 
\end{equation}
The hidden state of last character contains the global features of the input sentence, i.e., $\textbf{g} = \textbf{h}^c_n$.


\subsection{Lexicon Knowledge Encoding Layer}

To represent the semantic information of SLK of $i$th character, 
we embed $j$th lexical word ${{sw}}_{ij}$ in $\mathcal{SW}_i$ to distributional space as a semantic vector:
\begin{equation}
    \textbf{x}^{{sw}}_{ij} = \textbf{e}^w({{sw}}_{ij}),
\end{equation}
where $\textbf{e}^w$ is a pre-trained word embedding lookup table. 


\subsection{Contextual Lexicon Knowledge Fusion}
Not all lexical words contribute equally to the representation of the character meaning.
Hence, we introduce a global contextual information to 
extract such SLK
that are important to the meaning of the character 
and aggregate them to refine a character
vector.
Specifically, for the $j$th word in the matching set $\mathcal{SW}_i$ of the $i$th character,
we can obtain a hidden representation $\textbf{u}_{ij}$
for word embedding $\textbf{x}^{sw}_{ij}$:
\begin{equation}
    \textbf{u}_{ij} = \textbf{W}_u \textbf{x}^{{sw}}_{ij} + \textbf{b}_u,
\end{equation}
where $\textbf{W}_u$ and $\textbf{b}_u$ are update parameters. 
We measure the importance of lexical word as the similarity 
and get a normalized importance weight $\alpha_{ij}$.
Then, 
the SLK of $i$th character can be computed as
a weighted sum of the word information.
\begin{equation}
    \alpha_{ij} = \frac{exp(\textbf{u}_{ij}^{T} \textbf{g})} 
    {\sum_j exp(\textbf{u}_{ij}^{T} \textbf{g})},
\end{equation}

\begin{equation}
    \textbf{h}^{{sw}}_i = \sum_j \alpha_{ij} \textbf{x}^{{sw}}_{ij}.
\end{equation}


Finally, the final representation of $i$th character is denoted as $\textbf{r}_i = [\textbf{h}^{sw}_i; \textbf{h}^c_i]$.

\subsection{Decoding and Training}
To formulate the dependencies between successive labels, a standard CRF layer
is used to make sequence tagging.
We define matrix $\textbf{O}$ to be scores calculated based on the 
final representations $\textbf{R}=\{ \textbf{r}_1,...,\textbf{r}_n \}$: 
\begin{equation}
\textbf{O} = \textbf{W}_o \textbf{R} + \textbf{b}_o,
\end{equation}
where $\textbf{W}_o$ and $\textbf{b}_o$ are trainable parameters.
Then, the probability of tag sequence $y=\{ y_1,...,y_n\}$ is:
\begin{equation}
    p(y|s) =  \frac{
        exp(\sum_{i}(
            \textbf{O}_{i,y_i}
            +\textbf{T}_{  y_{i-1},y_i  }))
        }
        {\sum_{\hat{y}} exp(\sum_i 
        \textbf{O}_{i,\hat{y}_i}
        +\textbf{T}_{\hat{y}_{i-1},\hat{y}_i}))},
 \end{equation}
where $\textbf{T}$ is a transition score matrix,
and $\hat{y}$ denotes all possible tag sequences.
While decoding, we apply the Viterbi \cite{viterbi1967error} algorithm to get label sequence with the highest score.

Given training examples $\{ (s_j, y_j )\} |^N_{j=1}$, we optimize the model by minimizing
the negative log-likelihood loss:
\begin{equation}
    L = - \sum_{j} log(p(y_j|s_j)).
\end{equation}


\section{Experiments}
\subsection{Experimental Settings}
\subsubsection{Datasets}
As shown in Table~\ref{datasets}, we evaluate our model on three datasets, \textbf{OntoNotes4}, \textbf{Weibo} and \textbf{Resume}. OntoNotes4 is a multilingual corpus in the news domain
 that contains four types of named entities. 
{Weibo} 
dataset consists of annotated NER messages drawn 
from Sina Weibo\footnote{https://www.weibo.com}.
The corpus contains PER, ORG, GEP, and LOC for both named entity and nominal mention. 
 {Resume} dataset is composed of resumes collected from Sina Finance\footnote{https://finance.sina.com.cn/stock/}.
It is annotated with 8 types of named entities.
For OntoNotes4, we use the same training, validing and testing splits as \cite{DBLP:conf/naacl/CheWML13}. Since other datasets have already been split, we don't change them. 

\begin{table}[h]
    \caption{The statistics of the datasets.}\label{datasets}
    \centering
    \begin{tabular}{|l|c|c|c|c|}
    \hline
    \textbf{Dataset} & \textbf{Training} & \textbf{Validation} & \textbf{Testing}
         \\
    \hline
    OntoNotes4& 15724& 4301& 4346     \\
    Weibo & 1350  & 270       & 270  \\
    Resume   & 3821      &  463        &             477             \\
    \hline
    \end{tabular}
    \end{table}

    \begin{table*}[h]
        \caption{Experimental results(\%) on three datasets.
         }\label{result1}
        \centering
        \begin{tabular}{
            |l|c|c|c|
            c|c|c|
            c|c|c|}
        \hline
        \multicolumn{1}{|c|}{\multirow{2}{*}{\textbf{Models}}} 
        & \multicolumn{3}{c|}{\textbf{OntoNotes4}}  
        & \multicolumn{3}{c|}{\textbf{Weibo}}  
        & \multicolumn{3}{c|}{\textbf{Resume}}       
        \\
        \cline{2-10}  
           
        & P & R & F1
        & P &R & F1
        & P & R & F1
        \\
        \hline
         BiLSTM-CRF\cite{huang2015bidirectional}  
            & 72.0	& 75.1	& 73.5
            & 60.8	& 52.9	& 56.6
            & 93.7	& 93.3	& 93.5
            \\ 
        BERT\cite{devlin-etal-2019-bert}        
            &\textbf{78.0}	& 80.4	& 79.2     
            &61.2	            & 63.9	& 62.5 
            &94.2	            & 95.8	& 95.0
            \\
        CAN\cite{zhu2019can}  
            &75.1   &72.3   &73.6
            &55.4   &63.0   &59.3.
            &95.1   &94.8   &94.9
            \\
        \hline
        LGN\cite{gui2019lexicon}                
            & 76.1	& 73.7	& 74.9
            & {-}	&{-} 	& 60.2
            & 95.3	& 95.5	& 95.4 
            \\
        MG-GNN\cite{ding2019neural}   
            &  74.3	            & 76.2	& 75.2
            & \textbf{63.1}	& 56.3	& 59.5
            & {-}    &{-}    &  {-}   
            \\
        CGN\cite{sui2019leverage}  
            &75.1 &74.5 &74.8
            &-   &  -   &63.1
            &{-}    &{-}    &  {-}  
            \\
        \hline
        LatticeLSTM\cite{zhang2018chinese}
            &  76.4	& 71.6	& 73.9
            &  53.0	& 62.3	& 58.8
            &  94.8	& 94.1	& 94.5
            \\
        WC-LSTM\cite{liu2019encoding}  
            & 76.1	& 72.9	& 74.4
            & 52.6	& \textbf{67.4}	& 59.8
            & 95.3	& 95.2	& 95.2
            \\
        LR-CNN\cite{gui2019cnn}
            & 76.4	            & 72.6	& 74.5
            & -	                &-  	& 59.9  
            &\textbf{95.4}	&94.8	& 95.1
            \\ 
         
        \hline
        
        \textbf{SLK-NER} 
            & 77.9	&\textbf{82.2}	&\textbf{80.2}
            & 61.8	& 66.3              &\textbf{64.0}
            & 95.2	&\textbf{96.4}	&\textbf{95.8}
            \\
        \hline
        \end{tabular}
        \end{table*}

    \begin{table*}[htbp]
            \caption{Experimental results (\%) of different encoding strategies on three datasets.}\label{result2}
                \centering
            \begin{tabular} {|l|
                c|c|c|
                c|c|c|
                c|c|c|
                }
            \hline
            \multicolumn{1}{|c|}{\multirow{2}{*}{\textbf{Encoding Strategy}}} 
            & \multicolumn{3}{|c|}{\textbf{OntoNotes4}}  
            & \multicolumn{3}{|c|}{\textbf{Weibo}}  
            &\multicolumn{3}{|c|}{\textbf{Resume}}       
            \\
            \cline{2-10}
            \multicolumn{1}{|c|}{}  
            & \multicolumn{1}{|c|}{ P }  & \multicolumn{1}{|c|}{ R } & \multicolumn{1}{|c|}{ F1 } 
            & \multicolumn{1}{|c|}{ P } & \multicolumn{1}{|c|}{ R } & \multicolumn{1}{|c|}{ F1 } 
            & \multicolumn{1}{|c|}{ P } & \multicolumn{1}{|c|}{ R } & \multicolumn{1}{|c|}{ F1 } 
             \\
            \hline
            \textbf{using SLK} 
                & \textbf{ 77.9 } 	  &82.2 	&\textbf{ 80.2 }
                & \textbf{ 61.8 }     &66.3 	&\textbf{ 64.0 }
                & \textbf{ 95.2 }     &\textbf{ 96.4 } 	&\textbf{ 95.8 }  
                \\
            using FLK
                &76.6 	       &\textbf{ 82.9 } 	&79.8 
                &\textbf{ 61.8 } &	64.6 	&63.2 
                &95.1 	       &96.2 	&95.6   
                \\
            {using SLK and FLK} 
                & 76.4  	  &82.7 	&79.6
                & 60.6 &63.6&62.1
                &94.9 & 96.2 &95.5  
                \\
            no lexicon
                &77.7 	&81.3 	&79.6 
                &56.7 	&\textbf{ 66.5 } 	&61.2 
                & 94.2 	&96.1 	&95.1
                \\
            \hline
            \end{tabular}
            \end{table*}

    \begin{table*}[htbp]
                \caption{Experimental results (\%)  of different fusion strategies on three datasets.
            }\label{result3}
                \centering
                \begin{tabular}{|l|
                    c|c|c|
                    c|c|c|
                    c|c|c|
                    }
                    \hline
                    \multicolumn{1}{|c|}{\multirow{2}{*}{\textbf{Fusion Strategy}}} 
                    & \multicolumn{3}{|c|}{\textbf{OntoNotes4}}  
                    & \multicolumn{3}{|c|}{\textbf{Weibo}}  
                    &\multicolumn{3}{|c|}{\textbf{Resume}}       
                    \\
                    \cline{2-10}
                    \multicolumn{1}{|c|}{}  
                    & \multicolumn{1}{|c|}{P}  & \multicolumn{1}{|c|}{R} & \multicolumn{1}{|c|}{F1} 
                    & \multicolumn{1}{|c|}{P} & \multicolumn{1}{|c|}{R} & \multicolumn{1}{|c|}{F1} 
                    & \multicolumn{1}{|c|}{P} & \multicolumn{1}{|c|}{R} & \multicolumn{1}{|c|}{F1} 
                     \\
                    \hline
                    \textbf{Global-Attention} 
                        &77.9	&\textbf{ 82.2 } 	&\textbf{ 80.2 } 
                        &\textbf{ 61.8 }	&\textbf{ 66.3 } 	&\textbf{ 64.0 }
                        &\textbf{ 95.2 } 	&\textbf{ 96.4 } 	&\textbf{ 95.8 } 
                        \\
                    Self-Attention
                        &77.2 	&81.2 	&79.1 
                        &55.9 	&60.1 	&57.9 
                        &94.2 	&96.3 &	95.2 
                        \\
                    Shortest Word First
                        &77.1   &81.5   &79.2  
                        &55.8 	&57.7 	&56.7   
                        &93.9 	&96.1 	&95.0   
                        \\
                    Longest Word First 
                        &77.1     &81.6  & 79.3   
                        &57.6 	&56.9 	&57.3 
                        &94.7 	&96.1 	&95.4 
                        \\
                    Average 
                        &\textbf{  78.6 }    &80.8     & 79.7     
                        &56.4 	&58.4 	&57.3  
                        &94.3 	&96.3 	&95.3 
            
                        \\
                    \hline
                    \end{tabular}
            \end{table*}

\subsubsection{Comparisons} 
The methods evaluated are as follows.
    \textbf{BiLSTM-CRF} \cite{huang2015bidirectional}  was a sequence labeling model consisting of BiLSTM layer and CRF layer.
    \textbf{BERT} \cite{devlin-etal-2019-bert} was a pre-trained model with deep bidirectional transformer.
    \textbf{CAN} \cite{zhu2019can} investigated CNN-based model with attention layers to capture features of the character and its contexts.
    \textbf{Lattice-LSTM} \cite{zhang2018chinese} encoded characters in a sequence and all potential words that match a lexicon.
     \textbf{LGN} \cite{gui2019lexicon} used lexicon to construct the graph and provide word-level features.
    The literature \cite{ding2019neural} applied a multi-digraph structure to incorporate gazetteer information, and we denote \textbf{MG-GNN} for convenience. 
    \textbf{WC-LSTM} \cite{liu2019encoding} was used 
    to add word information into the start or the end character of the word. 
    \textbf{LR-CNN} \cite{gui2019cnn}  extended the rethinking mechanism when using lexicon.
     \textbf{CGN} \cite{sui2019leverage} investigated collaborative graph network (CGN) to leverage lexical knowledge.

\subsubsection{Implementation Details}
We use lexicon and word embeddings provided by \cite{li-etal-2018-analogical},
which is pretrained on Chinese Giga-Word 
using word2vec model. 
For character embeddings, we apply the {bert{-}base Chinese} model\footnote{https://github.com/google-research/bert} 
(12-layer, 768-hidden, 12-heads). 
For characters and words that do not appear in the pretrained embeddings, we initialize them with a uniform distribution\footnote{The range is $[-\sqrt{\frac{3}{dim}}, +\sqrt{\frac{3}{dim}}]$, where $dim$ denotes
the size of embedding.}.
When training, character embeddings and word embeddings are updated along with other parameters.
For hyper-parameter configuration, 
we set max length of sentences to 250, word embedding size to 50, 
the dimensionality of Bi-GRU to 512, the number of Bi-GRU layer to 1, 
the dropout to 0.1,
the batch size to 32.
We use Adam to optimize all the trainable parameters with learning rate $5e-5$. 
For evaluation,  we use the Precision(P), Recall(R) and F1 score(F1) as metrics in our experiments.

\subsection{Experimental Results}



Firstly, we compare SLK-NER with three general sequence labeling model for NER. 
All of them performed without any lexicon knowledge.
The results in the first block in Table ~\ref{result1},
show that our proposed model achieves best F1 and R,
which proves the efficiency of SLK-NER. 

Next, the second block in Table~\ref{result1} shows the performance of graph-based models.
SLK-NER gives better F1 and R than LGN, MG-GNN and CGN.
Although these baselines explore lexicon knowledge via the graph structure,
they performed without the consideration of contextual information.
Hence, we attribute the benefits to the efficiency of global context-aware in SLK-NER. 

Furthermore, the third block in Table ~\ref{result1} shows results of state-of-the-art sequence-based models.
We can observe that our proposed model achieves a remarkably improvement on F1 over three datasets. 
The results strongly verify the integrating SLK into character-based model 
enables to boost the performance.
By leveraging the SLK properly, 
our model is capable of improving NER in various domains, such as social network, news and Chinese resume.

\subsection{Strategies Analysis}
In this part, we explore the effects of strategies about lexicon knowledge.

\subsubsection{Lexicon Knowledge Types}
We conduct comparative experiments on different kinds of lexicon knowledge.
The results are illustrated in Table ~\ref{result2}.
We can clearly see that the character-based model performs poorly without lexicon knowledge,
demonstrating the usefulness of lexicon.
Besides, adding FLK makes a small improvement on F1.
While adding SLK outperforms significantly on F1 in all datasets.
The fact demonstrates the efficiency of SLK,
and reveals that leveraging second-order lexicon knowledge can indeed alleviate the word boundary conflicts.
Interestingly, when using both FLK and SLK, the F1 declines over three datasets. 
We conjecture the reason is there may be
some negative word conflicts simultaneously for a character
which limit the performance.

\subsubsection{Lexicon Knowledge Encoding}
We analyze the difference between the strategy in our model (Global-Attention) with four strategies proposed by \cite{liu2019encoding} 
for encoding word information, including Self-Attention, Shortest Word First, Longest Word First and Average.
The results in Table ~\ref{result3} show that global attention in our model achieves best performance on F1 score.
This demonstrates that our model can combine more informative features to determine the word boundary
and effectively alleviate the negative influence of word boundary conflicts.

\subsection{Sentence Length Analysis}

\begin{figure}[h]
    \centering{\includegraphics[width=0.35\textwidth]{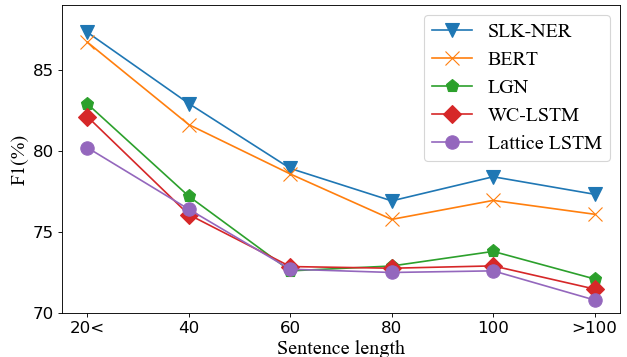}}
    \caption{F1 against sentence length on OntoNotes4 dataset. We split samples into six parts according to the sentence length.}
    \label{length}
\end{figure}

Fig.\ref{length} shows the F1 score of several baselines and SLK-NER against
sentence length on OntoNotes4 dataset. 
BERT and SLK-NER outperform significantly than other baselines, which indicates 
the ability to capture long dependencies. However, BERT ignores the word information among the sentence.
SLK-NER obtains a higher F1 over different sentence lengths compared to BERT,
which proves the SLK and global context-aware can capture more useful contextual information.

\section{Conclusion}
In this paper, we have investigated a lexicon-based model
  in Chinese NER task.
We present a new insight about second-order lexicon knowledge 
to incorporate informative lexicon into character-based model.
Based on this insight, SLK-NER is proposed 
to integrate more contextual word information into each character utilizing the global context. 
SLK-NER can effectively alleviate the impact of word boundary conflicts and 
word segmentation errors.
Extensive experiments on three public datasets have demonstrated 
 the superior performance of SLK-NER than state-of-the-art models.

\end{document}